\documentclass[letterpaper,10pt,conference]{ieeeconf}
  \IEEEoverridecommandlockouts{}
  \usepackage{amsmath,amssymb,amsfonts,bm}
  \usepackage{siunitx}
  \usepackage{graphicx}
  \usepackage{xcolor}
  \usepackage{booktabs}
  \usepackage{balance}
  \usepackage[many]{tcolorbox}
  \usepackage{blindtext}
  \usepackage[hidelinks,bookmarksopen=true]{hyperref}
  \usepackage[all]{hypcap}
  \usepackage{cite}
  \usepackage[nameinlink,capitalise]{cleveref}

\crefformat{equation}{(#2#1#3)}
\crefrangeformat{equation}{(#3#1#4) to~(#5#2#6)}
\crefmultiformat{equation}{(#2#1#3)}%
{ and~(#2#1#3)}{, (#2#1#3)}{ and~(#2#1#3)}

  \DeclareUnicodeCharacter{2212}{-}

  \usepackage{footnote}
  
  \def\vec#1{\boldsymbol{#1}}
  \def\set#1{\{#1\}}
  \def\tuple#1{(#1)}

  \def\qty#1#2{\SI{#1}{#2}}
  \def\qtyproduct#1#2{\SI{#1}{#2}}
  \def\num#1{#1}
  \def\numproduct#1{\num{#1}}
  \def\pct#1{\qty{#1}{\percent}}

  \definecolor{mplC0}{HTML}{1F77B4}
  \definecolor{mplC1}{HTML}{FF7F0E}
  \definecolor{mplC2}{HTML}{2CA02C}
  \definecolor{mplC3}{HTML}{D62728}
  \definecolor{mplC4}{HTML}{9467BD}
  \definecolor{mplC5}{HTML}{8C564B}
  \definecolor{mplC6}{HTML}{E377C2}
  \definecolor{mplC7}{HTML}{7F7F7F}
  \definecolor{mplC8}{HTML}{BCBD22}
  \definecolor{mplC9}{HTML}{17BECF}

  \def\FloorGenT{FloorGenT}

  \def\tymes{{\times}}

  \def\eg{e.g.}
  \def\ie{i.e.}
  \def\cf{Cf.\space}

  \def\p{p}
  \def\[#1\]{\begin{equation}#1\end{equation}}
  \def\eqnend#1{}
  \def\TokenSet{\mathcal{T}}

\begin{document}


  \def\t{\FloorGenT{}: Generative Vector Graphic Model of Floor Plans for Robotics}
  \title\t

  \author{Ludvig Ericson, Patric Jensfelt%
  \thanks{[Placeholder] Paper accepted for the 99th Earth Conference on Papers
  (ECP 3000). All authors are with the Division of Robotics, Perception and
  Learning at KTH Royal Institute of Technology, Stockholm, SE-10044, Sweden.
  This work was supported by the the VR grant XPLORE3D. For e-mail
  correspondence, contact {\tt\small{ludv@kth.se}}. \newline
  123-1-1234-1234-1/12/\$12.34 \textcopyright{} 1234 ABCD.
  }
  }

  \hypersetup{
    pdftitle=\t,
    pdfauthor={Ludvig Ericson, Patric Jensfelt},
    pdfcreator=,
    bookmarksopenlevel=2
  }

  \maketitle

  \def\kthds{KTH floor plan dataset}

\begin{abstract}

  Floor plans are the basis of reasoning in and communicating about indoor
  environments. In this paper, we show that by modelling floor plans as
  sequences of line segments seen from a particular point of view, recent
  advances in autoregressive sequence modelling can be leveraged to model and
  predict floor plans. The line segments are canonicalized and translated to
  sequence of tokens and an attention-based neural network is used to fit a
  one-step distribution over next tokens. We fit the network to sequences
  derived from a set of large-scale floor plans, and demonstrate the
  capabilities of the model in four scenarios: novel floor plan generation,
  completion of partially observed floor plans, generation of floor plans from
  simulated sensor data, and finally, the applicability of a floor plan model
  in predicting the shortest distance with partial knowledge of the
  environment.


\end{abstract}

\usetikzlibrary{fit,shapes,calc,positioning}

\section{Introduction}

  Reasoning and planning in indoor environments is an important
  capability in the context of robotics. One approach to this
  problem is through modelling floor plans. Floor plans are a simplified and
  minimalist map of an indoor environment, and can be used for both reasoning
  and communicating about such environments. In this paper, the aim is to
  predict floor plans based on environmental cues in order to facilitate
  robotics tasks in unknown environments, such as autonomous exploration and
  search-and-rescue. To this end, we construct a generative model that encodes
  the implicit rules of floor plans without explicitly stating those rules,
  \eg, that walls join at right angles, rooms are symmetrical, and doorways
  join rooms.

  \begin{figure}[t]
    \centering
    \includegraphics[width=0.95\linewidth]{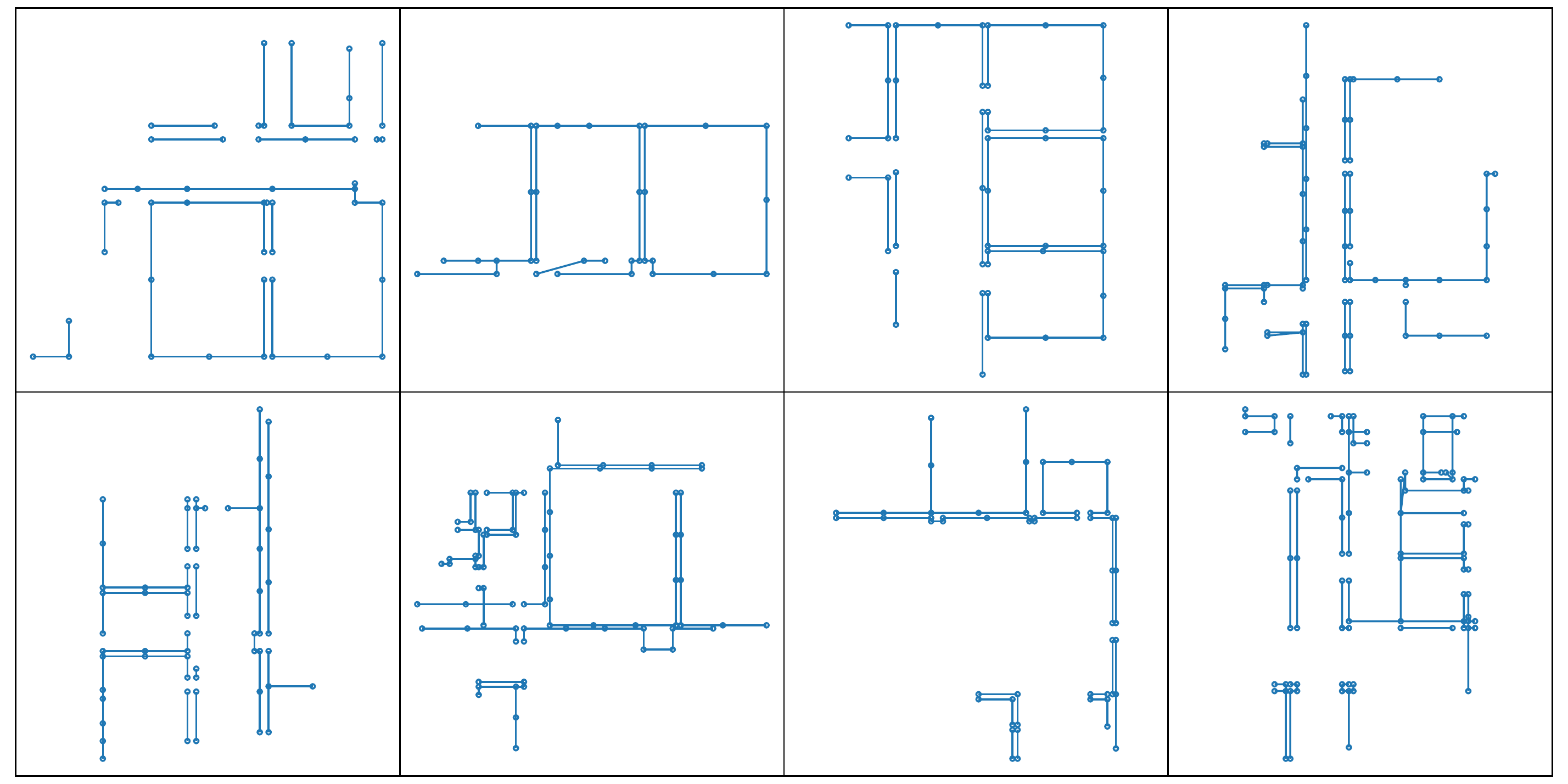}
    \par{\vspace{0.5ex}}
    \includegraphics[width=0.95\linewidth]{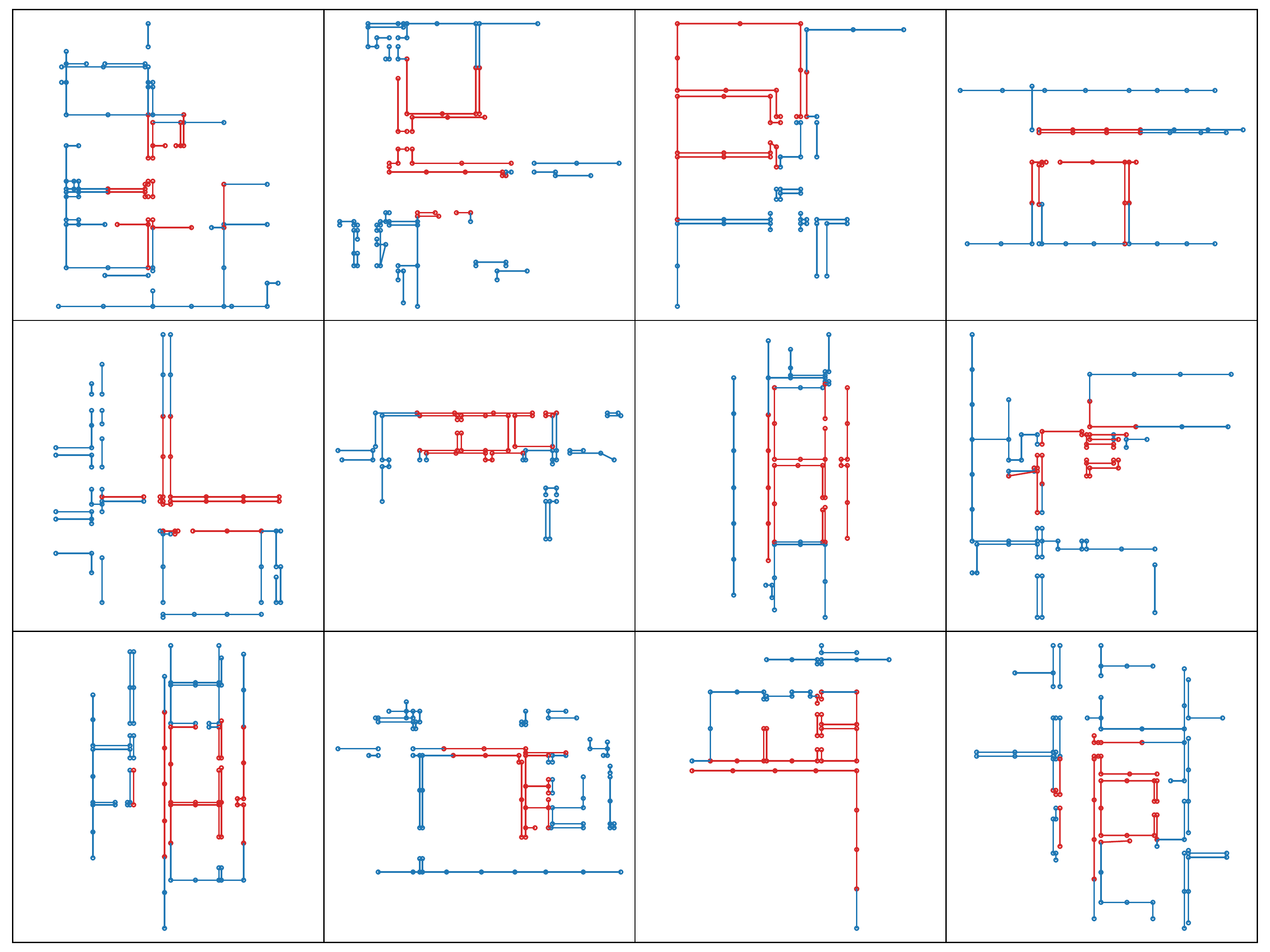}
    \par{\vspace{0.5ex}}
    \includegraphics[width=0.95\linewidth]{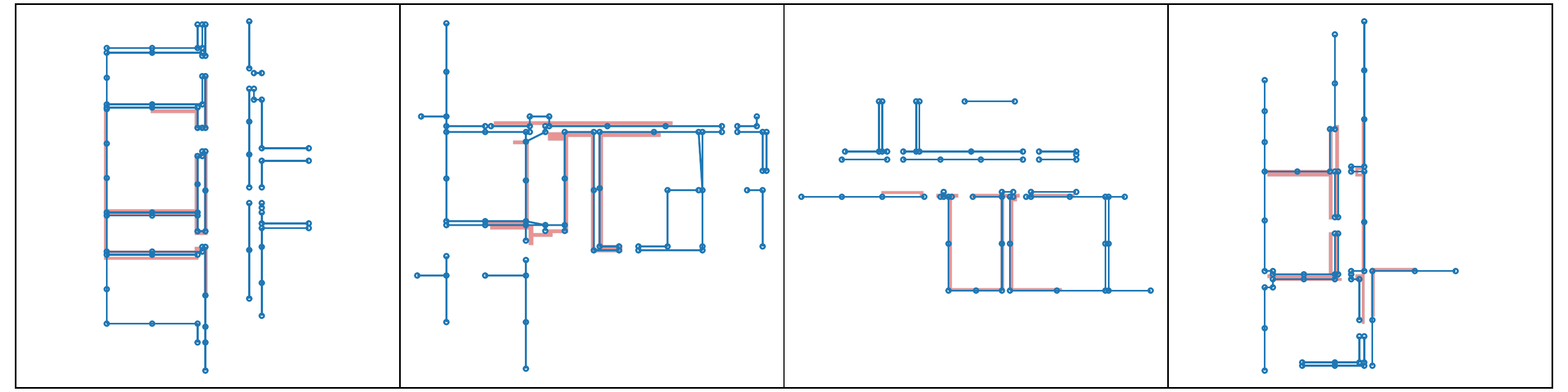}
    \caption{Randomly picked samples of a \FloorGenT{} network trained on the
    \kthds{} generated with nucleus sampling ($p=\pct{90}$). Blue is sampled
    model output. {\bf Top}: unconditioned novel samples. {\bf Middle}: partial
    sequence completion samples conditioned on the segments shown in red (first
    \num{25} segments of randomly selected test sequences, i.e., novel data to
    the network). {\bf Bottom}: partial image conditioned samples with the
    input image shown in red (rasterization of the first \num{25} segments of
    randomly selected test sequences).}%
    \label{fig:samples}
  \end{figure}

  Predicting floor plans from environmental cues is not a novel idea, and it
  has previously been cast as an image in-painting problem where an image
  model, typically a convolutional neural network, is used to fill in the
  missing regions of a rasterized floor plan. However, convolutions imply an
  inductive bias towards local dependencies in pixel space which is not
  necessarily a good fit for floor plans, which often exhibit non-local
  correlations. Furthermore, floor plan rasterizations are unlike natural
  images consisting exclusively of high-frequency content, making them
  particularly prone to artifacts such as bleeding and blurring. Care must be
  taken to prevent such effects as shown in~\cite{zeng2019deep}.

  \def\citeA{\cite{oord2016pixel,gpt2}}

  By modelling floor plans as vector graphics, this issue is avoided. It also
  lets us leverage recent advances in language models, such as attention-based
  neural network architectures. In an autoregressive sequence formulation such
  as~\citeA{}, the model takes its own previous output into account at each
  step as it generates an output sequence. This approach has shown remarkable
  results in modelling and generating various sequences, including but not
  limited to natural languages. By casting floor plans as sequences, similar
  performance could be obtained in the floor plan domain, and leveraged in a
  robotics context.

  In this paper we present such an implicit model of indoor environments, and
  demonstrate its usefulness in a typical robotics tasks. In summary, the
  contributions of this paper are:
  \begin{enumerate}
    \item A method of representing floor plans as a sequence of ordered line
          segments seen from a particular point of view within the floor plan,
          and a procedure to canonicalize the sequences.
    \item A attention-based generative model tailored to dealing with such
          sequences of line segments.
    \item An evaluation of the performance in generating diverse novel floor
          plan sequences, sequence completions from partial sequences, and
          sequence generation from partial birds-eye view raster images.
    \item A demonstration of the model's abilities in solving a typical
          robotics task, namely path planning in a partially observed
          environment.
  \end{enumerate}


  \def\githublink{\href{https://github.com/lericson/FloorGenT/}{\nolinkurl{github.com/lericson/FloorGenT/}}}
  \def\demolink{\href{https://lericson.se/floorgent/demo/}{\nolinkurl{lericson.se/floorgent/demo/}}}
  \def\spicylink#1{\mbox{{\ttfamily{}\bfseries{}\footnotesize{}#1}}}

  The network architecture and related code is accessible at \spicylink{\githublink{}}, and an
  online demonstration is available at \spicylink{\demolink{}}.

\section{Related Work}

  Floor plans as the basis of robotic planning and reasoning has been explored
  before. In the category of predicting maps, approaches range from explicit to
  implicit. A recent example of an explicit method is
  \cite{luperto2019predicting}, in which an algorithm for predicting the
  layouts of partially observed rooms in indoor environments is proposed, by
  constructing an \emph{explicit} set of rules and assumptions about those
  environments in the algorithm itself. A majority of recent work is based on
  deep learning, learning the rules from data rather than by design from an
  expert. The present work belongs to this latter \emph{implicit} category, but
  differs in a key regard: previous work is largely based on the convolutional
  neural networks from computer vision, such as
  \cite{saroya2020online,wang2020learning}. We instead cast the problem as
  modelling a sequence of line drawing instructions.
  
  Contrastingly, \cite{aydemir} use a \emph{topological} rather than metric
  representation, \eg, ``offices connect to kitchens through corridors'', which
  is then used for symbolic reasoning and planning. In pursuit of this, the
  \emph{\kthds{}} was constructed, consisting vector graphics of \num{38000}
  rooms in \num{184} floor plans from \num{27} different university buildings.
  We use the same dataset, though not for its topological information. It is
  important to note that these floor plans are large, with hundreds of rooms
  per floor.

  Separately from the robotics-aimed line of work is what is becoming known as
  \emph{inverse CAD}, where the goal is mapping images or pointclouds to
  architectural blueprints. For example, \cite{chen2019floorsp} propose a
  method to produce a vector graphics floor plan from a set of RGBD sensor
  scans that fully cover an apartment. In~\cite{zeng2019deep}, a method is
  proposed to classify room types (\eg{}, kitchen, bathroom, bedroom) from
  rasterized floor plan drawings. Both are based on convolutional neural
  networks. Inverse CAD is in general aimed at realtors and architects, rather
  than robotics.


  Though not di\-rect\-ed at solv\-ing robotics tasks, recent work has also
  been aimed at modelling vector graphics generally. For example,
  \cite{carlier2020deepsvg,reddy2021im2vec} show that vector graphics can be
  embedded in a latent space of fixed dimensionality. The embeddings then have
  some spatial relationship to each other, such that similar drawings are
  nearby in that space. Neither work is aimed at prediction from partial
  inputs.

  In terms of network architecture, we leverage recent developments in sequence
  modelling for natural languages in a similar style to attention-based models
  such as \cite{bert,gpt2}. In \cite{polygen}, a method is proposed to model 3D
  meshes as sets of polygons through the use of attention-based sequence models
  as proposed in \cite{aiayn}. Our model is built on that same foundation.



  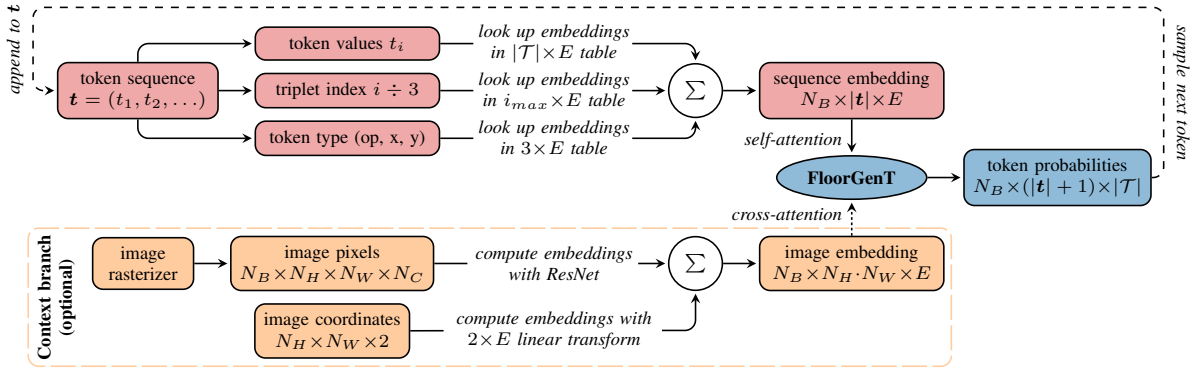
\begin{figure*}[ht]
    \centering

\colorlet{dsquery}{mplC3!40}
\colorlet{dscontext}{mplC1!40}
\colorlet{dsfloorgent}{mplC0!50}

\begin{tikzpicture}[
  > = stealth,%
  shorten > = 1pt, 
  auto,%
  node distance=1ex and 3ex,
  semithick, 
  scale=1,%
  align=center,%
  rounded corners,
  font=\scriptsize,
  tensor/.append style={rectangle,rounded corners,draw=black,font=\scriptsize},
  mapping/.append style={fill=none,font=\itshape{}\scriptsize},
  query/.append style={fill=dsquery},
  context/.append style={fill=dscontext},
  floorgent/.append style={ellipse,font=\bfseries{}\scriptsize,fill=dsfloorgent},
  fps/.append style={ellipse,fill=dsfps},
]

  \node[query, tensor, text width=14ex] (tokvals) {token values $t_i$};
  \node[query, mapping, inner xsep=0, text width=13ex, right=of tokvals] (tokvalembedding) {look up embeddings in $|\TokenSet| \tymes E$ table};

  \node[query, tensor, text width=14ex, below=of tokvals] (tokinds) {triplet index $i \div 3$};
  \node[query, mapping, inner xsep=0, text width=13ex, at=(tokvalembedding |- tokinds)] (tokindembedding) {look up embeddings in $i_{max} \tymes E$ table};

  \node[query, tensor, text width=12ex, left=of tokinds] (tokseq) {token sequence $\vec{t} = (t_1, t_2, \ldots)$};

  \node[query, tensor, text width=14ex, below=of tokinds] (tokdims) {token type (op, x, y)};
  \node[query, mapping, inner xsep=0, text width=13ex, at=(tokindembedding |- tokdims)] (tokdimembedding) {look up embeddings in $3 \tymes E$ table};

  \node[query, mapping, circle, draw=black, right=of tokindembedding] (sum1) {$\sum$};
  \node[query, tensor, text width=14ex, right=of sum1] (seqemb) {sequence embedding $ N_B \tymes |\vec{t}| \tymes E$};

  \node[floorgent, draw=black, below=3ex of seqemb] (floorgent) {\FloorGenT{}};

  \node[floorgent, tensor, right=of floorgent] (tokprobs) {token probabilities \\ $ N_B \tymes (|\vec{t}|+1) \tymes |\TokenSet| $};

  \node[context,tensor, text width=14ex, below=3ex of floorgent] (pixelsemb) {image embedding $ N_B \tymes N_H{\cdot}N_W \tymes E$};

  \node[context, mapping, circle, draw=black, at=(sum1 |- pixelsemb)] (sum2) {$\sum$};

  \node[context, mapping, inner xsep=0, text width=14ex, at=(sum2 -| tokvalembedding)] (pixelembedding) {compute embeddings with ResNet};
  \node[context, tensor, left=of pixelembedding] (pixels) {image pixels \\ $ N_B\tymes{}N_H \tymes N_W \tymes N_C$};

  \node[context, mapping, inner xsep=0, text width=18ex, below=of pixelembedding] (pixcoordembedding) {compute embeddings with $2\tymes{}E$ linear transform};
  \node[context, tensor, below=of pixels] (pixcoords) {image coordinates \\ $ N_H \tymes N_W \tymes 2$};

  \node[context, tensor, draw=black, text width=7ex, left=of pixels] (imraster) {image rasterizer};


  \node[fit=(imraster) (pixcoords) (pixels) (pixelsemb),inner sep=.5ex] (inner) {};
  \node[rotate=90,above,text width=11ex,font=\scriptsize] at (inner.west) (innerlab) {\bf Context branch (optional)};
  \node[draw=dscontext,semithick, dash pattern=on 9.5pt off 2pt, inner xsep=0,inner ysep=-0.5ex,fit=(inner) (innerlab)] {};

  \draw[->] (tokseq) |- (tokvals);
  \draw[->] (tokseq) to (tokinds);
  \draw[->] (tokseq) |- (tokdims);
  \draw[-] (tokvals) to (tokvalembedding);
  \draw[-] (tokinds) to (tokindembedding);
  \draw[-] (tokdims) to (tokdimembedding);
  \draw[->] (tokvalembedding) -| (sum1);
  \draw[->] (tokindembedding) -- (sum1);
  \draw[->] (tokdimembedding) -| (sum1);
  \draw[->] (sum1) to (seqemb);
  \draw[->] (seqemb) -- (floorgent) node[mapping,pos=0.6,left] {self-attention};;
  \draw[->, dash pattern=on 1pt off 1pt] (pixelsemb) -- (floorgent) node[mapping,pos=0.6,left] {cross-attention};
  \draw[->] (floorgent) to (tokprobs);
  \draw[->]  (imraster) to (pixels);
  \draw[-]  (pixels) to (pixelembedding);
  \draw[-] (pixcoords) to (pixcoordembedding);
  \draw[->] (pixelembedding) to (sum2);
  \draw[->] (pixcoordembedding) -| (sum2);
  \draw[->] (sum2) to (pixelsemb);

  \coordinate (col south east) at ($ (tokprobs.east) + (1ex,0) $);
  \coordinate (col north west) at ($ (tokseq.west) + (-2ex,7ex )$);
  \coordinate (col north east) at (col south east |- col north west);
  \coordinate (col south west) at (col north west |- tokseq.west);
  \draw[->, dashed] (tokprobs.east)
                 -- (col south east)
                 -- (col north east)
                 node[mapping,rotate=-90,above,midway,align=center] {sample next token}
                 -- (col north west)
                 -- (col south west)
                 node[mapping,rotate=90,above,midway,align=center] {append to $\vec{t}$}
                 -- (tokseq.west);

\end{tikzpicture}
    \caption{Overview of data flow in \FloorGenT{}. The input sequence is a
    possibly empty sequence of tokens $\vec{t}$, where each token is embedded
    as a sum of three discrete embedding vectors, is the input to the first
    self-attention layer. For the image models, the embedded input image is
    input to the cross-attention layers. When sampling, the next token is
    repeatedly drawn from the next token distribution, and fed back into the
    network at the end of the token sequence.
    }\label{fig:dataflow}

  \end{figure*}

\section{\FloorGenT{}}\label{sec:floorgent}


  \def\floorplan{\mathcal{F}}
  \def\OpEOS{{\mathtt{stop}}}
  \def\OpMove{{\mathtt{move}}}
  \def\OpLine{{\mathtt{line}}}

  %
  We model the distribution over floor plans $\floorplan$ by translating them
  to sequences of line drawing instructions, encoded as a \emph{token sequence}
  $\vec{t}$ and factorizing in autoregressive manner: \[
    \label{eq:factorization}
    \p(\floorplan) \triangleq \p(\vec{t}) = \prod_{i=1}^k \p(t_i|\vec{t}_{<i})
  \] where $t_i$ is the $i$th token in the sequence, and $\vec{t}_{<i}$ are the
  tokens before $i$. The tokens combine into triplets $\tuple{c, x, y}$ to form
  the line drawing instructions, consisting of the opcode $c \in{}
  \set{\OpEOS{}, \OpMove{}, \OpLine{}}$ meaning ``end of sequence'', ``move to
  coordinate'', and  ``draw line to coordinate'' respectively, and the operands
  are the discrete coordinates $x,y \in{} \set{q_1, q_2, \ldots, q_{N_Q}}$.
  We denote the set of tokens \[
    \label{eq:vocab}
    \TokenSet = \set{\OpEOS{}, \OpMove{}, \OpLine{}, q_1, q_2, \ldots, q_{N_Q}}\eqnend{.}
  \] 


  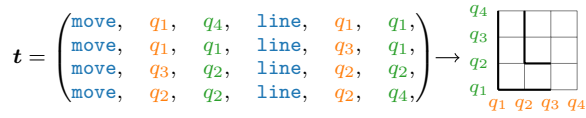
\begin{figure}[htb]
    \colorlet{tokop}{mplC0}
    \colorlet{tokx}{mplC1}
    \colorlet{toky}{mplC2}
    \def\trip#1#2#3{{\color{tokop}#1}, & {\color{tokx}#2}, & {\color{toky}#3}}
    \def\ls#1#2#3#4{\trip{\OpMove}{#1}{#2}, & \trip{\OpLine}{#3}{#4}}
    \centering{
    \begin{tikzpicture}[scale=0.35,font=\footnotesize,node distance=2ex]
      \draw[very thin,gray,font=\scriptsize]
        (1, 1) grid (4,4)
        (1, 1) node[below,black] (x1) {\color{tokx}$q_1$}
        (2, 1) node[below,black] (x2) {\color{tokx}$q_2$}
        (3, 1) node[below,black] (x3) {\color{tokx}$q_3$}
        (4, 1) node[below,black] (x4) {\color{tokx}$q_4$}
        (1, 1) node[ left,black] (y1) {\color{toky}$q_1$}
        (1, 2) node[ left,black] (y2) {\color{toky}$q_2$}
        (1, 3) node[ left,black] (y3) {\color{toky}$q_3$}
        (1, 4) node[ left,black] (y4) {\color{toky}$q_4$}
        ;
      \draw[thick] (1, 4) -- (1, 1)
                   (1, 1) -- (3, 1)
                   (3, 2) -- (2, 2)
                   (2, 2) -- (2, 4)
                   ;
      \node[fit=(y1) (y4) (x1) (x4),inner sep=0] (container) {};
      \node[anchor=east, inner sep=0, left=of container] (tokseq) {$
        \vec{t} = \begin{pmatrix}
          \ls{q_1}{q_4}{q_1}{q_1}, \\
          \ls{q_1}{q_1}{q_3}{q_1}, \\
          \ls{q_3}{q_2}{q_2}{q_2}, \\
          \ls{q_2}{q_2}{q_2}{q_4}, \\
        \end{pmatrix}
      $};
      \draw[->] (tokseq) -- (container);
    \end{tikzpicture}
    }
    \caption{An example of a token sequence and its corresponding drawing in
    the shape of an L. Note that in practice, the line segments would be sorted
    by their distance to some origin coordinate as described in
    \cref{sec:canonicalize}.}%
    \label{fig:example}
  \end{figure}

  Each line segment is encoded as two triplets, a $\OpMove$ followed by a
  $\OpLine$, illustrated in \cref{fig:example}. This representation allows
  manipulating sequences at line segment level, with the drawback that it is
  redundant when a line segment is joined with the previous segment. This is
  not typical in our case, as about \pct{12.4} segment pairs are so joined.

  \subsection{Token Sequence Model}

    \def\Attention#1{\operatorname{Attn}#1}
    \def\softmax#1{\operatorname{softmax}#1}

    With the sequence representation defined, we turn to modelling the
    distribution over sequences defined in \cref{eq:factorization}. We use the
    Transformer decoder architecture with \emph{multi-head scaled dot product
    attention} as in \cite{aiayn}. These networks were originally proposed to
    model natural languages, and our token sequences can similarly be thought
    as a synthetic language with its semantics defined primarily by the
    resulting vector graphic, making it necessary to ``read between the
    lines''.

    \def\Weights{\theta}
    \def\Loss{\mathcal{L}}

    The architecture is presented in \cref{sec:netarch}. The output of the model
    is a categorical distribution $\p(t_i|\vec{t}_{<i}; \Weights)$, where
    $\Weights$ denotes the parameters of the model. The loss $\Loss$ is the
    negative log likelihood of the ground truth token values $\hat{t}$ under
    the model's induced distribution, \[
      \label{eq:loss}
      \Loss(\hat{t}, \Weights) = \sum_i \log \p(t_i=\hat{t}_i \,|\,
      \vec{t}_{<i}=\hat{\vec{t}}_{<i}; \Weights)\eqnend{.}
    \]

  \subsection{Embeddings}

    We use discrete embedding vectors for the token sequence as illustrated in
    \cref{fig:dataflow}. Though it is possible to embed each vertex
    $\tuple{x,y}$ with a single token by summing or concatenating each
    coordinate's embedding, \cite{polygen} reported that this reduced
    performance significantly, and we therefore keep them embedded as separate
    tokens. It is also possible to remove the opcode tokens, however,
    \cite{weiss2021thinking} suggest that separator tokens allow the network to
    better propagate information between attention heads.


    Some variants of our model take images as inputs, and these are embedded by
    feeding the pixel values through an off-the-shelf convolutional neural
    network, together with pixel coordinate embeddings from a learned linear
    transformation of the pixel coordinates. This is illustrated as an optional
    context branch in \cref{fig:dataflow}. We evaluate two types of image
    embedding networks: first, as in~\cite{polygen}, a pre-activation style
    ResNet~\cite{he2016identity}, and second, the more recently proposed MLP
    Mixer~\cite{mlpmixer}. Neither network is pretrained.

  \subsection{Turning Floor Plans into Token Sequences}

    In the \kthds{}, a floor plan $\floorplan$ is defined as a set of
    \emph{spaces}, with each space having a bounding polygon defined as a
    sequence of line segments. Each segment has a type, \eg, wall, window, or
    ``portal'', \ie, a connection to another space. For our purposes, we treat
    windows as walls, and portals are ignored.

    We model an \emph{in situ} agent partially observing an environment with
    some sensor, \eg, a LIDAR, at some location in the environment. This lets
    us target on-line estimation scenarios such as autonomous exploration or
    search-and-rescue. It also means the dataset is ``over-sampled'', in the
    sense that we generate many training samples from a single floor plan.

    \def\P{\mathcal{P}}
    \def\S{\mathcal{S}}
    \def\dpmin{d_{pmin}}

    We sample \emph{valid} locations inside a floor plan by rejection sampling
    $N_P$ points $\P$ by sampling uniformly over the floor plan's bounding box
    until each sample are inside a space by some minimum clearance. We then
    maximize the distance between the \emph{selected} points $\S \subset \P$ by
    iteratively constructing $\S_i = \S_{i-1} \cup \set{\vec{s}_i}$ from the
    selected point $\vec{s}_i$ at each iteration where \[
      \vec{s}_i = \operatorname*{\arg\max}_{\vec{p} \, \in \, \P \, \setminus \, \S_{i-1}}
      \Big( \min_{\vec{p}' \, \in \, \S_{i-1}} {\|\vec{p} - \vec{p}'\|}_2 \Big)
    \] until $|\vec{s}_i - \vec{s}_{i-1}| < \dpmin$. We initialize with $\S_1 =
    \set{\vec{p}_1}$ without loss of generality as all $\vec{p}_i$ are
    independent and identically distributed. 

  \subsection{Canonicalization}\label{sec:canonicalize}

    \def\Nsegs{N_{\mathit{segs}}}
    \def\Nraster{N_{\mathit{raster}}}
    \def\LS#1#2#3#4{from $\tuple{#1, #2}$ to $\tuple{#3, #4}$}

    To remove extraneous degrees of freedom, we canonicalize the line segments
    so the set of token sequences that produce identical visual result is as
    small as possible. First, let the canonical order of endpoints of a line
    segment \LS{x_0}{y_0}{x_1}{y_1} be such that $x_0 < x_1$, and $y_0 < y_1$
    in the case when $x_0 = x_1$. In other words, the segments are always
    oriented left-to-right, or bottom-to-top for vertical lines.

    The line segments are centered and scaled by a global scale factor,
    determined by the dataset statistics, before being quantized. We extend the
    segments by considering any two overlapping parallel lines, then extending
    them to each other's furthest endpoints, resulting in one longer line. This
    yields a representation where each unbroken line segment is represented by
    only one actual segment. We then break each line segments at each
    intersection with another segment, so that no segments cross in their
    interior. Finally, we subdivide each segment so the longest segment length
    is bounded.

  \subsection{Partial Sequences}

    For each sample location $s$, the corresponding floor plan's line segments are
    ordered by their distance to $s$, and only the first (nearest)
    $\Nsegs$ line segments are kept. The distance is computed as the Euclidean
    distance between the location and the closest point on the line segment.
    The segments are translated so that the point-of-view location is at
    the origin. Finally, only the $\Nsegs$ nearest segments are kept and mapped
    to a sequence of tokens by inserting the appropriate opcodes.

  \subsection{Image Rasterization}

    A line drawing algorithm is used to rasterize the first
    $\Nraster~\ll~\Nsegs$ segments into a black-and-white bitmap to approximate
    a partial birds-eye-view occupancy grid. We do not use antialiasing.

\section{Experiments}

  In this section, we present both quantitative metrics and qualitative
  demonstrations of the model's output in different settings. In all
  experiments, the quantization level $N_Q$ is \num{256}, the sequence is at
  most $N_{segs}$~=~100 segments, each segment is at most 2.5~m long, the
  nearest segment is at least \qty{40}{cm} away, and the furthest segment is at
  most 7.5~m away.

  \subsection{Network Architecture and Training Details}

    We use six attention layers, $E$ is 512, and $N_{heads}$ is 8. See
    \cref{sec:netarch}. The models are trained with the Adam~\cite{adam}
    optimizer on the \kthds{}, divided into a 90--10 split with a learning rate
    of $3\cdot{}10^{-4}$. The dropout rate was set to \pct{60} for all models.
    Each model was trained over \num{e6} batches of size $N_B$~=~8, after which
    the optimization has converged. All layers use ReZero~\cite{rezero}, this
    accelerates convergence and allows deeper network architectures.

    \subsubsection{Data Augmentation}

      Each floor plans is sampled at high density with multiple overlapping
      sequences as a form of data augmentation, which also preserves invariants
      imposed by the canonicalization such as segment order. We then mirror and
      rotate by a multiple of \ang{90} at random using a \emph{signed
      permutation}. There are exactly eight such signed permutation matrices,
      generated by enumerating all combinations of three primitive operations:
      mirror X axis, mirror Y axis, and swap X and Y axes. 

  \def\ResNet{ResNet}
  \def\MLPMixer{MLP Mixer}

  %

  \begin{table}[tb]
    \vskip 2mm
    \caption{Evaluation and comparison of selected models}%
    \label{tab:eval}
    \centering
    \begin{tabular}{lccc} 
    \toprule
    Model & NLL (bits) & Top-1 & Top-5 \\
    \midrule
    Uniform                                     & 8.02 & \hphantom{\pct{12.3}}\llap{\pct{0.4}} & \hphantom{\pct{12.3}}\llap{\pct{1.9}} \\
    Nearest Neighbor                            &  ||  & \pct{59.7} & \pct{63.3} \\
    Equivalent MLP                              & 1.87 & \pct{68.2} & \pct{84.3} \\
    \midrule
                       \bf \FloorGenT{}         & 1.09 & \pct{81.4} & \pct{91.8} \\
    \hphantom{+}\llap{-\,} Opcode Tokens        & 1.12 & \pct{81.0} & \pct{91.5} \\
    \hphantom{+}\llap{-\,} Position Embed.      & 1.34 & \pct{77.8} & \pct{90.0} \\
    \midrule
                       \bf \FloorGenT{} Images  & 0.83 & \pct{84.3} & \pct{95.0} \\
    \hphantom{+}\llap{-\,} Position Embed.      & 0.99 & \pct{81.7} & \pct{93.5} \\
    \hphantom{+}\llap{+}   All Segm., \ResNet{}  & 0.50 & \pct{89.5} & \pct{98.2} \\
    \hphantom{+}\llap{+}   All Segm., \MLPMixer{} & 0.40 & \pct{91.4} & \pct{98.9} \\
    \bottomrule
    \end{tabular}
  \end{table}

    \subsubsection{Train-Test Split}

      \def\thefootnote{\fnsymbol{footnote}}

      To avoid test data leaking into the training set, we split training and
      test before shuffling the set of sampled sequences. This ensures that a
      given floor plan is only in one of the splits. Since a building's floors
      can be similar or even identical, care must also be taken to ensure that
      a given \emph{building} is only in one of the splits. In our case, the
      floor plans are ordered by the building they belong to, so such leakage
      is prevented by the same mechanism.

  \subsection{Predictive Performance on Test Set}

    The \kthds{} test set contains \num{2.46} million tokens with an average
    sequence length of 363 tokens. Likelihood and accuracy metrics of the
    models and variants described below are presented in \cref{tab:eval}.
    Negative log likelihood (NLL) is reported on the test set as mean bits per
    token. Top-$k$ accuracy is computed as the frequency with which the ground
    truth token value is in the $k$ most likely tokens predicted by each model,
    with sequences from the test set. Top-1 is equivalent to the maximum
    likelihood estimate.

    \subsubsection{Uniform}

      We report a worst case reference point with uniform probabilities. The
      NLL is $\log |\TokenSet|$, and the top-$k$ accuracy follows a hyper\-%
      geometric distribution.

    \subsubsection{Nearest Neighbors}

      Predictions are formed by finding the $N_k$ most similar subsequences in
      the training set, using the sliding window approach outlined in
      \cref{sec:netarchmlp} to construct subsequences of $N_w$ tokens. The most
      common following token is then the prediction. Sequence similarity is
      measured by Hamming distance, \ie, the number of unequal tokens. We
      report performance with $N_w=10$ and $N_k=32$, tuned by hyperparameter
      search and cross-validated. NLL is not reported as the model is
      deterministic.

    \subsubsection{Equivalent MLP}

      We ablate the attention mechanism entirely, turning the network into a
      multi-layer perceptron (MLP), refer to \cref{sec:netarchmlp} for
      architecture details. Notably, this uses nearly four times more network
      parameters. The MLP had the best performance of the baselines we
      measured, though not as good as its attention-based counterparts.

    \subsubsection{\FloorGenT{}}

      This is the standard formulation of the model as described in
      \cref{sec:floorgent}.

    \subsubsection{Opcode Tokens}

       We evaluate a model that is only given coordinate tokens and the
       $\OpEOS$ opcode token in its input. We let $p(t_i=\hat{t}_i)=1$ for the
       removed opcode tokens, since the parity of the triplet index is
       sufficient to determine where $\OpMove$ and $\OpLine$ opcodes should be
       inserted. We find that performance is largely unchanged from the
       standard formulation, though inference time is shorter since a third of
       the tokens have been removed.

    \subsubsection{Position Embeddings}

      Line segments are the same regardless of their position in the token
      sequence, so their representation should arguably also be position
      invariant. We follow \cite{lee2019set} and remove the position
      embeddings, so the model is unable to rely on statistics of the sequence
      position. We find that though this delayed overfitting somewhat, dropout
      was still necessary. The accuracy and NLL was unsurprisingly worse, as it
      is not possible to know which token was last (or anywhere) in the
      sequence without position embeddings, and it is therefore more difficult
      to predict what the next token will be --- though not impossible since
      the segments are ordered by distance from the origin.



  \subsection{Novel and Partial Sequence Completion}
    Generative language models are often used to complete sentences with
    contextually relevant completions. In the same way, we can complete partial
    floor plan sequences, generating plausible continuations of partially
    observed floor plans. We present a set of such samples from our model in
    \cref{fig:samples}.

    Sampling is performed iteratively in an autoregressive manner. Given a
    partial or empty sequence, the network returns a distribution over possible
    next tokens, a sample is drawn from this distribution, and then added to
    end of the sequence. This iteration continues until the sampled token is
    the $\OpEOS{}$ token, or the maximum iteration count $i_{m{}ax}$ is
    reached. This process is illustrated in \cref{fig:dataflow}. Nucleus
    sampling~\cite{nucleus} is applied with top-$p$ at \pct{90}, which works by
    moving probability mass from the ``unreliable tail'' of the token
    distribution to its \emph{nucleus}.

    \begin{figure}[bt]
      \vskip 2mm
      \centering
      \includegraphics[width=\linewidth]{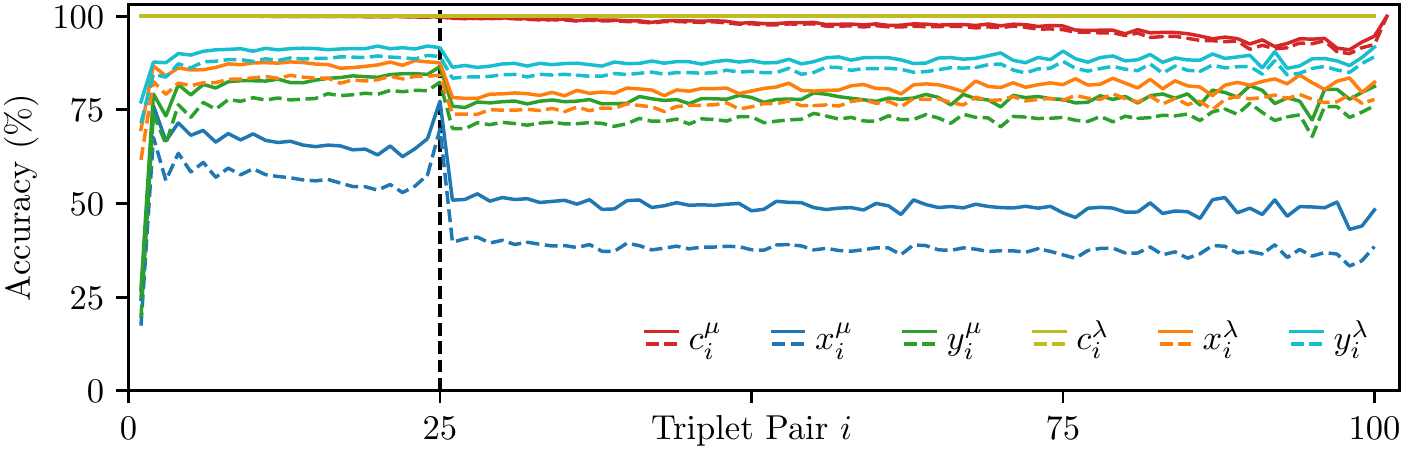}
      \caption{Predicted probability (dashed lines) versus empirical accuracy
      (solid lines) per token position on the test set. The six plots
      correspond to token types when grouped into $\OpMove{}$-$\OpLine{}$
      triplet pairs denoted $\mu$ and $\lambda$. Each point on the horizontal
      axis thus corresponds to a line segment. These results are from the
      partial image model, and the dashed vertical line separates segments
      visible in the input image from those that are not. The plot for
      $c_i^\mu$ extends one index longer since it includes the final $\OpEOS{}$
      token.}%
      \label{fig:dyadaccuracy}
    \end{figure}

  \subsection{Partial Image Conditioning}

    We condition the model on binary $128\tymes{}128$ pixel images by
    rasterizing the first 25 line segments. The image is intentionally made
    ambiguous: there a quarter as many pixels as there are quantized
    coordinates, and only a subset of the line segments are drawn into the
    image. The network must therefore both learn to reproduce line segments
    from a rendering of the 25 nearest, and generate up to 75 novel line
    segments that fit the first 25.

    To see the effect of the image embedding network, we report the performance
    of two models where \emph{all} line segments are rasterized, with two
    different networks, ResNets~\cite{he2016identity} and
    \MLPMixer{}~\cite{mlpmixer}. The \MLPMixer{} variant is \pct{40} smaller,
    yet it has \pct{20} lower NLL.
    The results are reported together with the non-image results in
    \cref{tab:eval}.

    The predicted token probability by the model, and the empirically estimated
    accuracy grouped by triplet pairs is presented in \cref{fig:dyadaccuracy},
    where $c^\mu, x^\mu, y^\mu, c^\lambda, x^\lambda, y^\lambda$ refer to a
    pair of $\OpMove{}$ and $\OpLine{}$ triplets (denoted $\mu$ and $\lambda$
    respectively). We find the predicted probability to always be slightly less
    than the accuracy, and the difference is more pronounced when the
    probability is lower. Both drop sharply after the 25th triplet pair, which
    corresponds to the last visible line segment. We find that $x^\mu$ more or
    less determines the other three coordinates, suggesting that polar
    coordinates $\tuple{\phi, r}$ may be easier to model since the line
    segments are mostly ordered by $r$.

  \subsection{Predicting the Shortest Path}

    \begin{figure}[t]
      \vskip 2mm
      \centering
      \includegraphics[width=\linewidth]{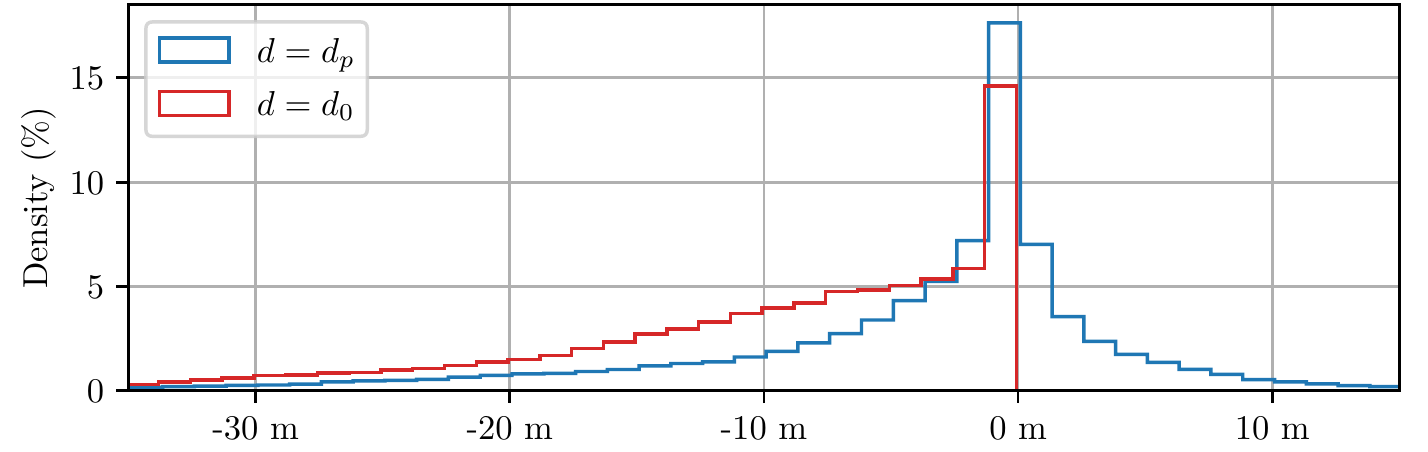}
      \caption{Histogram of the distance error $\epsilon=d-\hat{d}$ with
      distances estimated in a completed floor plans from the first 25 line
      segments of the test data ($d_p$), compared to distances estimated using
      just the 25 first line segments themselves ($d_0$). $\hat{d}$ is the
      distance using the true floor plan.   The histogram bins are eight cell
      sides wide. The two plots are shifted slightly horizontally to improve
      legibility.}%
      \label{fig:bfserror}
    \end{figure}

    \begin{figure}[b]
      \centering
      \includegraphics[width=\linewidth]{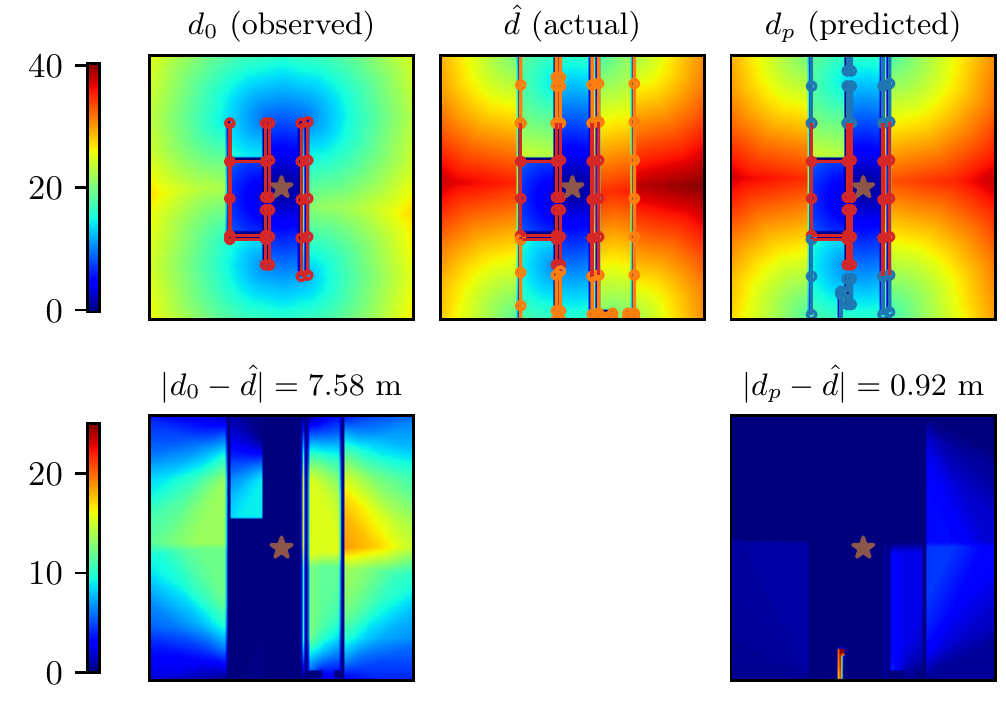}
      \caption{An example of the three distance grids $d_0, \hat{d}, d_p$, and
      the absolute error $|\epsilon|$ per cell. All quantities are in meters.
      The corresponding line segments are drawn on top of each grid. Red line
      segments are observed, blue predicted, and orange is the true floor plan.
      The star denotes the origin, \ie, from where the distance is computed.}%
      \label{fig:bfsexample}
    \end{figure}

    Finally, we evaluate the applicability of \FloorGenT{} to a typical
    robotics task, namely, predicting the distance (and path) to an unknown
    point in space given environmental cues. The scenario is as follows: a
    robot observes its immediate surroundings, precisely the 25 nearest line
    segments at a location in the test set. \FloorGenT{} is used to estimate
    the expected distance under the model given the observation via a Monte
    Carlo approximation. To that end, we generate 8 completions from the
    observation, and the completions are rasterized to create the occupancy
    grids $M_{p1}, \ldots, M_{p8}$. A shortest-path search from the origin to
    each cell in each occupancy grid is performed, with obstacles inflated by
    four cells. The predicted distance $d_p$ for each cell is then defined to
    be the median distance of the 8 completions. The same process is repeated
    for the ``null hypothesis'' occupancy grid $M_0$, created by rasterizing
    onyl the observed 25 line segments, and also repeated for the true floor
    plan rasterization $\hat{M}$. The occupancy grids are \numproduct{256x256}
    cells over an area of \qtyproduct{40x40}{m}. The search is 8-connected and
    uses Euclidean distance cost. We denote the distance under the null
    hypothesis $d_0$, and the distance in the true floor plan $\hat{d}$. A cell
    is considered trivial if $d_p=d_0=\hat{d}$ and is left out of the following
    statistical evaluation. An example calculation is presented in
    \cref{fig:bfsexample}.

    \def\EpsBase{\qty{-9.65}{m}}
    \def\EpsPred{\qty{-4.03}{m}}
    \def\AbsBase{\qty{9.65}{m}}
    \def\AbsPred{\qty{6.21}{m}}
    \def\AbsDiff{+\qty{3.44}{m}}

    The mean absolute error $|\epsilon|$ is \AbsPred{} using predictions and
    \AbsBase{} (\AbsDiff{}) under the null hypothesis, \ie, assuming that only
    the observed obstacles exist. A histogram of the errors is shown in
    \cref{fig:bfserror}. Note that the error $\epsilon$ is better centered
    around zero with predictions, while the null hypothesis by definition
    results in only underestimating, \ie, $\epsilon \le 0$ (mean $\epsilon$ is
    \EpsPred{} and \EpsBase{} respectively).

    Though we have only looked at the median distance, the same
    approach could for example be used to gauge certainty by computing the
    variance, or to assert the satisfaction of safety parameters, \eg, ``cell
    can be reached without running out of battery''.

\section{Conclusions}

  We have presented \FloorGenT{}, a generative model for floor plans, and
  showed its potency in modelling and predicting large-scale floor plans like
  the \kthds{}. We demonstrated the model's ability to incorporate other data
  modalities, in particular, rasterized images similar to occupancy grids which
  is a common representation in robotics applications, and showed its ability
  to reasoning about unknown space in a typical robotics task.

  We believe there are many interesting real-world use cases for predictive
  floor plan modelling, and particularly in conjunction with sensor data.
  Concretely, we plan to employ map predictions in an autonomous exploration
  scenario as proposed in our previous work \cite{ecmr21}. This could be
  accomplished with pretrained networks for the context embedding, and
  potentially combined with semi-supervised learning since there are relatively
  few datasets of indoor floor plans with sensory data, but abundant data from
  indoor scenes.



\bibliographystyle{IEEEtran}
\bibliography{your}

\appendix
\subsection{Network Architecture Details}%
\label{sec:netarch}

    \def\Tensor#1{\vec{#1}}
    \def\MultiheadAttn{\operatorname{MHA}}
    \def\LayerNorm#1{\overline{#1}}
    \def\Dense{\operatorname{Dense}}
    \def\ReLU{\operatorname{ReLU}}
    \def\Linear{\operatorname{Linear}}
    \def\MLP{\operatorname{MLP}}
    \def\AttnLayer{\operatorname{AttnLayer}}
    \def\Dropout{\operatorname{D}_r}
    \def\Nheads{N_{\mathit{heads}}}

    An attention layer is defined as $\Tensor{y}=\AttnLayer(\Tensor{x}_0)$ by
    \begin{align*}
      \Tensor{x}_1 &= \Tensor{x}_0 + \alpha_1 \Dropout(\MultiheadAttn(\LayerNorm{\Tensor{x}_0}, \LayerNorm{\Tensor{x}_0}; \Nheads, E)) \\
      \Tensor{x}_2 &= \Tensor{x}_1 + \alpha_2 \Dropout(\MultiheadAttn(\LayerNorm{\Tensor{x}_1}, \Tensor{v}_{\mathit{context}}; \Nheads, E)) \\
      \Tensor{y}   &= \Tensor{x}_2 + \alpha_3 \Dropout(\Linear(\Tensor{x}_{2}))
      \intertext{where $\Linear(\Tensor{z}_0) = \Tensor{z}_{2}$ with}
      \Tensor{z}_{1} &= \Dense(\LayerNorm{\Tensor{z}_0}; N_{fc}) \\
      \Tensor{z}_{2} &= \Dense(\ReLU(\Tensor{z}_{1}); E)\eqnend{.}
    \end{align*} The network is largely as proposed by \cite{aiayn}. $\Dropout$
    is dropout at rate $r$. $\LayerNorm{\Tensor{x}}$ is layer normalization as
    in~\cite{layernorm}. $\MultiheadAttn(\Tensor{q}, \Tensor{m})$ is multiple
    heads of scaled dot-product attention as in~\cite{aiayn}, with queries
    $\Tensor{q}$, keys and values $\Tensor{m}$. $\Dense$ is a fully-connected
    layer with given number of output units. $\alpha_i$ are the ReZero
    coefficients. $\Nheads$ is the number of attention heads. $E$ is the
    embedding dimension. $\Tensor{v}_{\mathit{context}}$ is the context
    embedding, \cf image embedding in \cref{fig:dataflow}. For non-image
    models, we let $\Tensor{x}_2 = \Tensor{x}_1$. Each successive layer
    operates on the output of the previous, so the output $\Tensor{y}_{net}$ of
    an $L$-layer network is defined \begin{align*}
      \Tensor{y}_{0} &= \textit{sequence embedding (\cf \cref{fig:dataflow})} \\
      \Tensor{y}_{i} &= \AttnLayer(\Tensor{y}_{i-1}) \\
      \Tensor{y}_{net} &= \LayerNorm{\Tensor{y}_{L}}
    \end{align*}
    Finally, $\Tensor{y}_{net}$ is projected to $|\TokenSet|$ dimensions to
    obtain the logits of the predictive distribution $\p(t_i|\vec{t}_{<i};
    \Weights)$.

    \subsection{Equivalent MLP}
    \label{sec:netarchmlp}

    \def\SlidingWindow{\operatorname{SlidingWindow}}
    \def\MLPLayer{\operatorname{MLP\,Layer}}
    \def\Join{\operatorname{Join}}
    \def\Split{\operatorname{Split}}
    \def\Reshape{\operatorname{RS}}
    \def\Nfc{N_{\mathit{fc}}}

    \balance

    In the ``Equivalent MLP'' model, we take $N_{w}$ sub\-sequences of each
    input sequence in a sliding window fashion, \eg, the sequence
    \textit{abcdef} would under a sliding window with $N_{w} = 3$ yield four
    subsequences: \textit{abc, bce, cde, def}. Padding items are prepended so
    that we obtain as many subsequences as there are tokens in the input, \eg,
    \textit{ssabcdef} with a start token \textit{s}. The network input
    $\Tensor{y}_{0}$ and output $\Tensor{y}_{net}$ is as before, but with the
    preprocessing step $\Tensor{y}_{1}$ and a different layer definition
    $\Tensor{y}_{i}$
    \begin{align*}
      \Tensor{y}_{1} &= \Join(\SlidingWindow(\Tensor{y}_{0}); N_{w}{\cdot}E) \\
      \Tensor{y}_{i} &= \MLPLayer(\Tensor{y}_{i-1})
    \end{align*}
    where $\SlidingWindow$ is said sliding window operation and yields a
    $S{\times}N_{w}{\times}E$ tensor for input length $S$. $\Join$ flattens the
    last two dimensions as indicated, \ie, a concatenation of the embeddings in
    each window. A single layer is defined $\Tensor{y} =
    \MLPLayer(\Tensor{x}_0)$ with
    \begin{align*}
      \Tensor{x}_1 &= \Dense(\LayerNorm{\Tensor{x}_0}; N_{w}{\cdot}\Nfc) \\
      \Tensor{x}_2 &= \Split(\Tensor{x}_1; N_{w}{\times}\Nfc) \\
      \Tensor{x}_3 &= \Dense(\ReLU(\Tensor{x}_1); E) \\
      \Tensor{x}_4 &= \Join(\Tensor{x}_3; N_{w}{\cdot}{}E) \\
      \Tensor{y}   &= \Tensor{x}_0 + \alpha \Dropout(\Tensor{x}_4)
    \end{align*}
    $\Split$ is the inverse of $\Join$ and unflattens the two last dimensions
    as indicated. The above setup mimics the attention layers: the first dense
    layer can ``attend'' to anything in the given window, and the second dense
    layer projects each position individually to $E$ dimensional embeddings. In
    our experiments, we have six MLP layers, $E=\Nfc=12$, and $r=\pct{5}$.

\end{document}